Optimistic Risk Perception in the Temporal Difference error Explains the Relation between Risk-taking, Gambling, Sensation-seeking and Low Fear

*Joost Broekens, Tim Baarslag*

*Interactive Intelligence*

*Delft University of Technology*

*The Netherlands*

**Abstract**

Understanding the affective, cognitive and behavioural processes involved in risk taking is essential for treatment and for setting environmental conditions to limit damage. Using Temporal Difference Reinforcement Learning (TDRL) we computationally investigated the effect of optimism in risk perception in a variety of goal-oriented tasks. Optimism in risk perception was studied by varying the calculation of the Temporal Difference error, i.e., *delta*, in three ways: realistic (stochastically correct), optimistic (assuming action control), and overly optimistic (assuming outcome control). We show that for the gambling task individuals with "healthy" perception of control, i.e., *action optimism*, do not develop gambling behaviour while individuals with "unhealthy" perception of control, i.e., *outcome optimism*, do. We show that high intensity of sensations and low levels of fear co-occur due to optimistic risk perception. We found that *overly* optimistic risk perception (outcome optimism) results in risk taking and in persistent gambling behaviour in addition to high intensity of sensations. We discuss how our results replicate risk-taking related phenomena.

**Introduction**

Understanding the affective, cognitive and behavioural processes involved in risk taking is essential for treatment and for setting environmental conditions to limit damage. Sensation-seeking, risk-taking behaviour, gambling disorder (pathological gambling), and (perception of) control over event outcomes positively correlate with each other [1-5]. Explanations for this co-occurrence include impulsiveness of the individual [6, 7], perceived self-efficacy [8], and optimistic risk perception. Optimistic risk perception results from event outcome probability bias [1, 8] characterized by exaggeration of positive outcome probabilities compared to negative ones. However, the mechanistic details responsible for the relation between sensation-seeking, risk-taking behaviour, gambling disorder, and (perception of) control remain unknown, in particular in the context of task-adaptive behaviour (learning).

Especially due to the complex nature of risk taking and gambling mechanisms, computational modelling has been argued to be a fruitful method of study [9], in particular to generate and test hypotheses about possible mechanisms. It is in this line of research that our contribution must be seen. We use Temporal Difference Reinforcement Learning (TDRL) learning [10-12], as a framework to investigate cognitive, affective and behavioural properties of risk taking. In this article we focus on one specific hypothesis: *the amount of optimism in risk perception can be modelled in the calculation of the Temporal Difference error, and this model can replicate the co-occurrence of low-fear, sensation intensity, risk-taking behaviour, gambling, and (perception of) control*.

Imagine you need to choose between two options: one option is a moderate reward; the other is a great reward with small probability and a small punishment with high probability. What do you pick? The answer depends on two things: the pay-off structure of the options, and, the weighting of the different outcomes and probabilities. Now imagine that you need to master an arbitrary choice task, i.e., neither the payoff structure of the options is given nor structure of the task. This involves learning what actions can be done in what situations, and what the consequences of these actions are as well as recursively modelling the utility of particular actions in particular situations. So to master this task, we learn by repetition to select actions that minimize negative future outcomes and maximize positive future outcomes over time for that task.

Let's have a look at risk perception in our example, and in particular the exaggeration of positive outcomes compared to negative ones. A gambler would ascribe too little importance to negative consequences of actions compared to a non-gambler. As a result, even after repeated trials, the gambler will take the risk of picking the gambling option, even if the outcome, on average, is lower than the moderate but certain reward of the other option. Gamblers take risk. This risk is not perceived as negative either, risk taking is associated with sensation-seeking [2] (thrill seeking behaviour). Further, risk-taking should be related to low fear: if you underestimate the gravity of negative outcomes by not considering them in the calculation of the utility of an action, then obviously you don't feel much fear either.

The question we address in this work is how a generic mechanism for learning goal directed tasks can replicate the co-occurrence of a diverse set of risk taking related phenomena including gambling behaviour, low-fear, perception of control, and sensation intensity. We want this single mechanism to explain not only this co-occurrence but also the learning of goal-oriented tasks in the first place. For example, no-one is a born gambler, while we are all born as individuals that are able to learn tasks from feedback and experience. In other words, we are all born with the capacity to learn complex behaviours that enable us to optimize expected return. As gambling is also a learned behaviour, this must mean that some of us are born with learning-related properties that make them prefer gambling situations. The question is: which learning-related properties explain not only gambling, but also the co-occurring phenomena including risk-taking, low-fear, perception of control, and sensation intensity.

To explain the learning and risk perception elements of this hypothesis intuitively, consider the example again. While you are learning to master the choice task, you update estimates of how good or bad a particular option is. Let's call this estimate the *state utility* of that option. Of course all states in the task, not only the options, will converge to a particular state utility. Assuming Temporal Difference Reinforcement Learning (TDRL) as a model for the way you would learn to master this task, the update signal you use to change a state's utility is called the *Temporal Difference error,* or *delta*. To calculate this signal you use the return received from the action executed now and the utility of the states that possibly follow this action. Your perception of risk is important for the calculation of the Temporal Difference error in the following way: an optimistic perception of risk puts too much emphasis on positive outcomes, will result in a signal that is consistently higher, and thus results in learning overvalued utilities for situations that involve potentially negative outcomes. This means that through repeated learning, you will perceive risky states to be too positive. In a similar fashion, if a person takes drugs that consistently boost the Temporal Difference error, then any and all behaviour related to and following drug administration will be sought for, simply because the signal is always positive due to the drug [10].

To explain the affective aspects of our hypothesis intuitively, consider how a gambler would feel if the great reward actually falls. This would involve a great deal of joy. Similarly, losing the gamble induces high levels of distress. We propose in this article that the intensity of the experience relates to the intensity of joy and distress, and that joy and distress are modelled by the Temporal Difference error. If it's positive, it's joy, if it's negative, it's distress. For individuals with a positive risk perception bias, this would mean that as long as there is any positive outcome possible, the amount of distress caused by a negative outcome is severely dampened. This must be the case as the negative outcome is not used that much in the calculation of the Temporal Difference error. In a similar way, a situation's utility is linked to hope and fear: positive utility models hope, negative utility models fear. These affective interpretations enable us to model sensation intensity (high joy/distress) and fear intensity in the context of learning a risky task. A finding that we were able to confirm in our experiments is, for example, that optimistic risk perception results in high intensity of joy, because the Temporal Difference error is consistently higher, providing an explanation for the relation between sensation intensity and gambling [3].

Overvaluing state utility due to exaggeration of probabilities of positive outcomes in and of itself is not a new idea, as it is the basis of utility-based risk perception, and in particular configural weight theory [13]. What is new is our integration of this idea with a computational framework that models goal-oriented learning. Others have linked addiction to Temporal Difference Reinforcement Learning [10, 14]. We extend this by explicitly incorporating affective aspects in the set of to be explained risk-taking related phenomena and we provide detailed simulations to do so.

In this article, we report on computer-simulated trials in which we investigated the effect of optimism in risk perception on a variety of symptoms related to risk taking. Each trial consists of learning to execute a goal-oriented task. The effect of optimism was studied by varying the calculation of the Temporal Difference error, i.e., the *delta*, used in learning to adapt to the task in three ways: realistic (stochastically correct), optimistic (assuming action control), and overly optimistic (assuming outcome control). Our results indicate that when the task models a gambling scenario, individuals with "healthy" perception of control, i.e., the assumption that one has control over ones actions (optimistic), do not develop gambling behaviour while individuals with "unhealthy" perception of control, i.e., the assumption that one has control over outcomes of actions (overly optimistic), do. In addition, we show that high intensity of sensations (intensity of joy and distress) and low levels of fear co-occur due to optimistic risk perception. Further, we found that overly optimistic risk perception results in risk taking and in persistent gambling behaviour in addition to high intensity of sensations. We discuss how our work computationally replicates many psychological findings including the finding that effects of perception of control on risk-taking are mediated by optimistic risk perceptions [1] and the positive correlation between sensation intensity and pathological gambling [15].

**Temporal Difference Reinforcement Learning.**

Reinforcement Learning [16] (RL) is a computational approach used for learning behaviour based on exploration and feedback and relies on a mechanism similar to operant conditioning. This approach is often used to allow artificial agents, autonomous pieces of software, to learn to solve a task. In Reinforcement Learning, the goal is to update the agent's *policy*, i.e., the way an agent selects actions, based on the experience of the agent such that the total reward received by the agent is maximized over the long run by that policy. In other words, RL is a computational method to solve the credit assignment problem: how much credit should an action *a* in state *s* get such that this credit is an estimate of the expected future return. For example, a robot that needs to learn to walk where the return is defined by the distance covered, or a computer program that needs to learn to play backgammon where the return is defined by winning or losing. More formally, Reinforcement Learning takes place in an environment with states *S*. An agent present in that environment selects an action *a∈A* to perform in a state *s∈S*. The action the agent executes is based on its policy *π*, with *π(a|s)* the probability that *a* is selected in *s*. Standard RL has two additional elements: the transition probability $P^a_{ss'}$, being the conditional probability *P(s'|as)* to end up in state *s'* when action *a* is executed in state *s*, and the reward $R^a_{ss'}$, being the reward (or punishment) the agent expects based on previous experience to receive for that state transition. To ensure that agents prefer shorter solutions rather than longer ones, a discount factor *γ*, with *γ<1* discounts rewards that are further in the future.

With these elements it is possible to define a value *V^π(s)* for each state (the state's utility). The values must be specific to the agent's policy *π* for selecting actions because values depend on the policy: e.g., agents that for some reason always pick random actions should learn to predict different values than agents that pick actions in a smarter way (hence the superscripted *π*). Now we can define the Bellman equation:

$$V^\pi(s) = \sum_{a \in A} \pi(a|s) \sum_{s' \in S} P^a_{ss'}(R^a_{ss'} + \gamma V^\pi(s')) \qquad (1)$$

This equation recursively defines the values of states as correct estimates of the to-be-expected future return, i.e., rewards followed by that state and the discounted values of states that follow next. "Correct" means "weighted average of", where weighing is based on the probabilities of selecting actions and the probabilities of ending up in particular states due to those actions.

The value of a state is typically arbitrarily initialized and updated as the state is visited more often. To update values, the simplest way is to use the Bellman equation (1) to update $V^{\pi}(s)$ each time a state is visited. This means that after each action $a$, $V^{\pi}(s)$ is recalculated to reflect the observed changes in its terms.

The next step for an agent is to use the learned values to select actions. Several types of action selection exist, ranging from completely random to greedy, i.e., choose the action resulting in maximum predicted value. The Boltzmann distribution (2), argued to be a model for human action selection [17], gives a probability $p(a)$ of choosing each action and can be scaled from acting as a greedy selection function to a random selection function and is given by:

$$p(a) = \frac{e^{\beta Q(s,a)}}{\sum_{b=1}^{n} e^{\beta Q(a,b)}} \qquad (2)$$

where $\beta$ is a positive parameter, called inverse temperature, defining the amount of randomness in the selection process, and $Q(s,a)$ is the value of taking a specific action according to:

$$Q(s,a) = \sum_{s' \in S} P_{ss'}^{a} (R_{ss'}^{a} + \gamma V^{\pi}(s')) \qquad (3)$$

Updating $V^{\pi}(s)$ or $Q(s,a)$ based on the Bellmann equation (1) requires building a model of the environment, as $P_{ss'}^{a}$ is needed. These probabilities are not always available. Temporal Difference Reinforcement Learning (TDRL) estimates values and updates them after each visit using a learning rate $\alpha$ to make sure values converge after sufficient state sampling. The simplest method, one-step TDRL, updates values according to:

$$V(s) \leftarrow V(s) + \alpha(r + \gamma V(s') - V(s)) \qquad (4)$$

with $\alpha$ representing the learning rate, and $(r + \gamma V(s') - V(s))$ the Temporal Difference error, i.e., *delta*. When TDRL is directly applied to learning action values instead of state values, updates are according to:

$$Q(s,a) \leftarrow Q(s,a) + \alpha(r + \gamma Q(s',a) - Q(s,a)) \qquad (5)$$

After convergence, action values *Q(s, a)* can be used to determine action preferences.

Reinforcement Learning, and TDRL in particular has been proposed as a plausible model for human and animal adaptation of behaviour based on feedback [11, 18, 19]. In RL an (artificial) organism learns, through experience, estimated utility of situated actions. It does so by solving the credit assignment problem, i.e., how to assign a value to an action in a particular state so that this value is predictive of the total expected reward (and punishment) that follows this action. After learning, the action selection process of the organism uses these learned situated action values to select actions that optimize reward (and minimize punishment) over time. In Reinforcement Learning, reward, action value (utility), and utility updates are *the* basic elements based on which action selection is influenced. These basic elements have been identified in the animal brain including the encoding of utility [20], changes in utility [21], and reward and motivational action value [20, 22-24]. These studies explicitly argue that these processes relate to instrumental conditioning, and in particular to Reinforcement Learning [25, 26].

**Joy, Distress, Hope and Fear in Reinforcement learning agents**

While most research on computational modelling of emotion is based on cognitive appraisal theory [27], computational studies show that affective signals can be conceptualized within the Reinforcement Learning framework. Reinforcement learning has been used to study the effect of simulated emotion on adaption [28-33], the effect of communicated emotions on robot learning [34], and to study how emotions emerge from reinforcement learning [35-37]. A recent proposal in the psychological literature explicitly reframed emotions in terms of feedback-based learning to device an alternative explanation of emotions as behaviour altering feedback signals as opposed to emotions as forces that directly cause behaviour [38]. Finally, a broadly agreed-upon function of emotion in humans and other animals is to provide a complex feedback signal for a(n) (synthetic) organism to adapt its behaviour [39-45]. While cognitive appraisal-based modelling provides a rich framework with respect to how the current situation is evaluated affectively (e.g., belief-desire-intention or planning-based simulations of complex emotions such as guild and anger are possible), the RL framework is more suitable for modelling how feedback mechanisms relate to affect and adaptive behaviour.

To investigate affective aspects of risk-taking behaviour, we need a computational measure for sensation intensity that is based on the TDRL framework. Sensation seeking is defined as "a trait defined by the seeking of varied, novel, complex, and intense sensations and experiences, and the willingness to take physical, social, legal, and financial risks for the sake of such experience" [46]. Sensation seeking is related to the amount of stimulation needed to maintain an optimal level of arousal [47]. There is no reference to the necessity of positive valence of the experience. Therefore we assume sensation intensity is related to the intensity of joy *and* distress. This means we need to model the intensity of joy and distress within the TDRL framework. We build upon earlier simulation work suggesting that the emotions of joy and distress can be modelled by the Temporal Difference error [36, 37]. Joy and distress are signals that habituate when the agent is presented repeatedly to the same reinforcement [48] and respond to hedonic (dis)pleasure but also anticipated (dis)pleasure [41, 49], and as such are signals that should respond to feedback and predicted feedback. The Temporal Difference error is such a signal, it habituates under repetition and responds to (changes in predicted) rewards and punishment or absence thereof. Other RL signals do not fare well in replicating habituation [36]. For example, reward never habituates, so joy cannot be the reward signal. This interpretation of joy is also consistent with most cognitive appraisal theories [41, 49], stating that joy is a signal that tells the individual that either something intrinsically pleasant has occurred or a goal has come closer. The responsiveness of joy to intrinsic pleasantness and goal-orientedness is nicely captured by the Temporal Difference error as that signal responds to reward and changes to anticipated future rewards. Finally, anticipated reward (e.g. in the form of a present) generates less joy than unanticipated reward, again nicely matching convergence of the Temporal Difference error. Our model for joy is consistent with the finding that dopamine activation in the ventral striatum is associated with Temporal Difference error calculation [11, 50] as well as with euphoria [51]. It is also a computational model of Rolls' [44, 52] proposition that a goal-directed habit produces less (or no) emotion intensity as opposed to the rewarded process responsible for learning the habit.

Further, we need a TDRL based model of fear. Fear and hope are signals that respond to anticipated goodness/badness. We build upon earlier simulation work suggesting that fear and hope can be modelled by means of *V(s)* [36, 37]. In cognitive appraisal theory, fear is an anticipation of negative future outcome [41] and hope is the anticipation of positive future outcome [41]. In hope theory [53] (and the closely related theory on optimism [54]) hope is seen as the tendency to see pathways towards goal achievement (i.e., a motivational state resulting from the successful execution of pathways leading towards goal achievement). High-hope individuals tend to explore pathways more effectively, while low hope individuals tend to get stuck in blocking thoughts [53]. One could say high-hope individuals see beams of light, while low-hope individuals see pools of mud. These views on hope and fear match with the state value *V(s)* of Reinforcement Learning. The state value represents *the* anticipation of future outcome, and is (at least in the model-based RL case) calculated based on the probability of pathways towards future rewards (goals). This view is not completely correct as it is possible to experience both hope *and* fear in the same state. As such, we propose that hope is the signal derived from positive pathways available in the current state *s*, i.e., $V_+(s)$, while fear is derived from negative pathways, i.e., $V_-(s)$. This is consistent with fear extinction, and in particular fear extinction due to new learning [55]. New associations *s – s''*, *s- s'''*, etc., other than a fear-conditioned *s - s'* (where s' is a punished state) reduce the contribution of the punishment to the value of state *s* due to the increasing probabilities of the new associations. Thus, the old association is not forgotten, but is made less important due to the new ones. This view of fear/hope is also consistent with hope disposition: high hope (optimistic) individuals overrate positive probabilities as compared to low-hope (pessimistic/realistic) individuals. Further, the emotions of hope and fear occur later in child development than joy and distress [56], which is per definition true for *V(s)* and *delta* respectively, as state values can only arise as a consequence of repeated updates. So, updates come before anticipation: joy/distress before hope/fear. This model for fear, $V_-(s)$, is compatible with the amygdala being essential for fear processing as well as for mediation of a gated (e.g., satiation dependent) association between stimuli and affective value [57, 58].

**Experiments**

We performed simulated trials to study the impact of risk perception on the intensity of sensations, the development of fear, the occurrence of risk-taking behaviour and the development of gambling behaviour. Individuals are modelled as computer programs that interact with an environment, called agents. In our case, the agents represent adaptive individuals and learn to adapt their behaviour using Reinforcement Learning [16].

*Independent factors*

We varied three levels of risk perception and four different tasks (3x4 setup). Each task (Figure 1a) computationally represents a variation of a to be learned choice problem. A task is a grid world with 9 or 11 states. A state is a discrete *<x,y>* location. The agent navigates through the task using the actions up, down, left and right. The state space is visualized as a T-maze with two rewarded options at the top.

The first task contains no risk but instead models a basic trade-off between a positive and a negative option. It has a left choice (A) that is rewarded (0.2) and the right choice (B) that is punished (-0.1). We refer to this task as *T-maze trade-off*.

The second task is the same but now the outcome of (B) is split in two states $B_1$ and $B_2$ defined as a biased coin with a payoff structure with $p(B_1)=0.9$ with feedback -0.2 and $p(B_2)=0.1$ with feedback

0.8, where *p(x)=y* defines the probability *y* to receive outcome *x*. Average payoff for choice (B) is thus equal to that in the T-maze trade-off task. This second task models a typical gambling scenario in which choice (A) (the distracter) has a small positive payoff and (B) (gambling) has a small average negative payoff characterized by frequent small losses and an occasional large win. We refer to this task as *gambling scenario.*

In the third task, reward is numerically represented by 0.2, while punishment is numerically represented by -0.1, as in the T-maze trade-off task. However, this time, when an individual receives feedback, that feedback is randomly placed at a non-occupied location picked from three possible feedback locations. This models risk introduced by unpredictable reward locations, in contrast to risk introduced by unpredictable outcomes such as in the *gambling scenario*. We refer to this task as *risky world*.

The fourth task uses the same structure as the T-maze trade-off task with the same payoff structure; however, individuals in this task have no control as actions are random and enforced upon the individuals. This models a situation in which the agent lacks control over selecting its actions in an otherwise predictable environment, i.e., one is able to predict consequences but can do nothing about it. We refer to this task as *lack of control.*

We computationally modelled optimism in risk perception as a bias in the calculation of the Temporal Difference error that an individual uses to update the expected payoff (the value) of visited states. While exploring the problem, each time an individual picks an action *a* in a state *s* (visualized as a location in the maze), and arrives in a next state *s'*, the new value for that state $V_{new}(s)$ is calculated according to *V(s')* and the feedback received in *s*, taking into account the odds of picking *a* and the odds of arriving in *s'*. The update $V_{new}(s) - V(s)$ is defined as *delta*, the Temporal Difference error [12, 25] for a particular state $s^1$. The calculation of *delta* thus depends on the estimation of $V_{new}(s)$ and hence on the probabilities of picking actions and the probabilities of the states that follow next. We defined three variations of risk perception.

---

[1] at a particular moment *t*. For readability we do not refer to the different exposures to *s* however.

In *realistic* risk perception, *delta* is calculated using the Bellman equation to get a correct estimate for *V(s)* based on observed probabilities for picking actions and resulting states, i.e., there is no bias. So, $V_{new}(s)$ is calculated at each visit of state *s*:

$$V_{new}(s) = \sum_a p(a|s) \sum_{s'} P^a_{ss'}(R^a_{ss'} + \gamma V^\pi(s')) \qquad (6)$$

Here, *p(a|s)* is the historical probability for that agent (as observed over learning) for selecting action *a*, with the other terms as defined above.

In *action-control biased* risk perception, *delta* is calculated using the standard *V\*(s)* (Max$_a$ off-policy) formula [16]. *V(s)* is updated based on the value of the best action available, disregarding the probability of selecting that action. This represents an *optimistic* individual who assumes full control over its actions, but not over the outcomes of those actions. In other words, $V_{new}(s)$ is calculated using the following formula:

$$V_{new}(s) = max_a \gamma \sum_{s'} P^a_{ss'}(R^a_{ss'} + \gamma V^\pi(s')) \qquad (7)$$

This update thus assumes a policy that selects the best possible action effectively setting the probabilities for selecting other actions to 0, i.e., it assumes the agent has control over its actions.

In *outcome-control biased* risk perception, *delta* is calculated based on the value of the best possible outcome *V\*\*(s)* (i.e., *Max$_{as'}$*). Now, the updates disregard the odds of *s'* actually happening (for example $s_3'$ in Figure 1b). This represents overly optimistic individuals in that they assume full control over the outcome of their actions. In other words, $V_{new}(s)$ is calculated using the following formula:

$$V_{new}(s) = max_{as'}(R^a_{ss'} + \gamma V(s')) \qquad (8)$$

Here, the new value for *V(s)* is updated only according to the best possible outcome, effectively setting the probabilities for selecting actions *and* for ending up in other states than *s'* to 0. This update thus (wrongfully) assumes that the policy can not only select the best possible action but also the outcome of that action, i.e., it assumes the agent has control over the environment. This might seem artificial (and stupid) from a Reinforcement Learning perspective, but it is in effect an extreme form of overestimating positive outcomes as opposed to underestimating negative outcomes. This is exactly the condition we want to test for the occurrence of risk-taking and gambling behaviour.

*Materials and simulated subjects*

For each of the 12 experimental conditions we sampled 5000 individuals from an agent population. In each condition, all individuals autonomously adapt their behaviour to the task independently of other individuals (i.e., there is no interaction between agents). In our case an individual moves through the states of the problem (locations) by selecting actions up, down, left or right. Each individual was given 5000 actions (one trial) to learn the task, after which it is deleted. Whenever an individual reaches a goal state (A or B), it is relocated to a randomly selected starting state (position in the maze) and then continues trying to solve the problem until the trial is over. The agent population from which samples are taken is defined by the range of possible values on RL-related parameters including discount factor, randomness in action selection and default action value. Parameters are fixed or varied with Gaussian noise. We set the discount factor $γ$ to 0.9 (with std=0.01), default action value to 0 (with std= 0.1), and $β$ in the Boltzmann action selection equal to 10(with std=1) (except in the *lack of control* setting where action selection is random and $β=0$)[2,3,4]. We used online learning for the agents, i.e., we do not use a separate learning phase in which the agent chooses actions randomly for a longer period of time to learn a correct model for *V(S)*. The agent learns and performs at the same time. Updates to *V(s')* influence action selection at the next visit of state *s*. Natural agents learn online: animals use earlier received feedback immediately the next time they encounter a similar situation, otherwise they risk dying.

---

[2] This is a typical value for $γ$ allowing the agent to learn longer sequences of actions but still have a clear discount of future rewards so that shorter solutions are preferred.
[3] Setting $β$ too high results in no exploration as action selection becomes greedy, while setting $β$ to 0 results in exploration but no exploitation of the learned model as action selection is random.
[4] Note that these parameter values and the simulation setup are standard and not the object of this study.

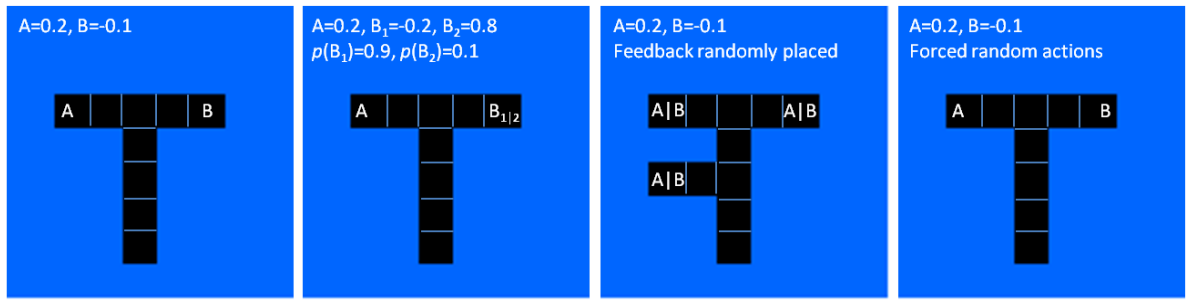

*Figure 1a. Grid-world Learning tasks: T-maze trade-off, Gambling scenario, Risky world, Lack of control. In the T-maze trade-off and lack of control tasks, the letters A and B refer a static payoff equal to 0.2 or -0.1; in the Gambling scenario $B_{1|2}$ is a probabilistic outcome with $p(B_1)=0.9$ and $p(B_2)=0.1$, with $B_1=-0.2$ and $B_2=0.8$; in the Risky world three goal locations are used of which only two contain payoff in on trial, and when the agent consumes the payoff, the payoff is randomly replaced at one of two empty locations, resulting in a random distribution of A and B over the three goal locations. In all task, the agent is randomly replaced at an empty location after receiving the payoff.*

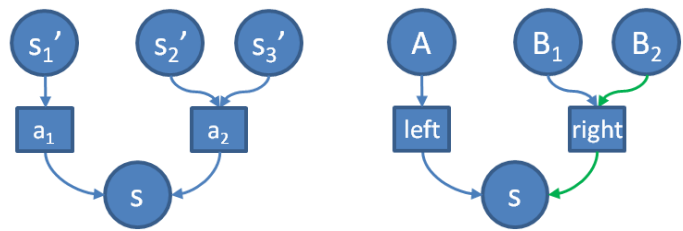

*Figure 1b. A typical s-a-s' Backup diagram (left) showing a deterministic outcome of action $a_1$ executed in state s, a probabilistic outcome of action $a_2$, and an instantiation of the Backup diagram for the gambling task (right) with a value update that assumes outcome control (green arrows).*

*Dependent measures*

To analyze the effect of task and risk perception variation, for each experimental condition we plotted average *fear intensity*, *sensation intensity* (in terms of joy and distress), and *reward* received over time. Fear is defined as the predicted future punishment in a particular state *s*, that is, $V_-(s)$. Sensation intensity is defined as a combination of joy, modelled by a positive Temporal Difference error (*joy = delta* if *delta* > 0), and distress, modelled by a negative Temporal Difference error

(*distress = delta* if *delta* < 0). Further, a negative average reward is a measure for *gambling behaviour* as the gambling option (B) has below zero pay-off. Finally, *Risk taking* is measured as the average number of times an individual chooses for the biased coin (B) in the gambling scenario.

**Discussion of results**

Our main results (Figure 2) show that control bias in risk perception (i.e., action-control *and* outcome-control taken together), as opposed to realistic risk perception, *in general* results in lower intensity of fear, lower intensity of distress, and higher intensity of joy in all learning tasks (see the figure captions for a reference to which RL signals joy, distress and fear refer to). For all tasks this is true in the early stages of learning (until trial 100). For the T-maze task there is no noticeable risk-perception-induced difference for fear, distress or joy over the long run. For the gambling scenario there is no noticeable difference for distress and joy over the long run, and a clear exception to this observation is that fear increases over the long run in the outcome-control bias case. For the risky world task the general rule is true over the long run, except for distress which is higher. In the lack of control task, the general rule is true for fear over the long run, but there is no noticeable difference for joy and distress. We first interpret the general observation after which we discuss the details of the exceptions task by task.

The general observation indicates that there is a clear affective benefit to the assumption of control. An extreme case that shows this benefit is the lack of control task, in which assuming control does not (in fact cannot) have an effect on actual learning outcomes (same rewards over time, same choice distribution between A and B), but it does result in a more pleasurable experience (higher joy, lower distress and fear) during the early stages of adapting to the task. This indicates that the perception of action control is not only useful for learning an appropriate value function in many cases [16], but also for coping with negative affect in situations where no control is possible, supporting the idea that the illusion of control can reduce stress [5] without harming performance. The general observation is also in line with hope theory [53] and optimism [54]: high hope individuals tend to see opportunities and possible pathways to their goals, while low hope individuals tend to see obstacles and think they are not capable of solving the problem. Our manipulation of risk perception directly manipulates the influence of positive versus negative "pathways" in valuing states, and replicates emotional consequences of optimism. Further, the general observation extends findings related to the association between perceived control, optimism, and risk perception [1] in the following way: optimism in risk perception is the responsible variable that can also explain affective qualities of optimism (i.e., a disposition to higher wellbeing and favourable expectancies [54]), and this could explain the relation between dispositional optimism and coping [59].

The straightforward explanation for the absence in the T-maze task of a risk-perception-induced difference for fear, distress or joy over the long run is that learning converges similarly in all three conditions. The task does not involve uncertainty of any kind (such as in the other three tasks), and as such it does not matter over the long run for the agent whether or not it assumes anything about this uncertainty in its value updates. The difference in the early learning phase is due to the optimistic risk perception in the Temporal Difference error which emphasizes the left branch of the T-maze (choice A). This emphasis results in higher Temporal Difference errors *and* state values that quickly move away from the occasional negative experience from the right branch of the T-maze (choice B) resulting in low fear and high joy. This result should be interpreted as follows: learning to adapt to a novel static situation that involves control over positive versus negative outcomes produces higher levels of well-being for those who have optimistic risk-perception. This is again in line with hope theory and optimism, as explained above [53, 54].

For the outcome-control biased gambling scenario we observed a strong preference for gambling with negative average pay-off compared to a positive pay-off distracter. This is in line with configural weight theory (for review see [13]): the subjective probability of the negative outcome $B_1$ of the gambling choice is completely disregarded, because $B_1$ and $B_2$ are both children at the same level in the tree with parent *s* and all the weight goes to the positive outcome $B_2$. In other words, more attention is given to the positive outcome resulting in less (in our extreme case no) attention to the negative outcome. The decaying average reward indicates that over time each individual eventually starts gambling: because the coin is a stochastic process and each agent always has a non-zero chance to explore the gambling choice due to the Boltzmann action selection, the moment at which the gambling choice takes over varies. Some agents get exposed to the positive gambling outcome early on, while others get exposed late. Because we plot *mean* reward, this is observable as decay. *Mean* joy remains small but present for quite some time reflecting the strong positive Temporal Difference error that percolates all the way back through previous states when individuals encounter a positive gambling pay-off ($B_2$) for the first time. This Temporal Difference error is strongly reduced in the realistic and action-control case because on average the positive effect of $B_2$ percolation through previous states is limited due to stochastic dominance of the negative outcome $B_1$. Joy and distress eventually converge to 0, also in the outcome-control case for the same reason as mentioned in the T-maze case: all learning converges; hence the Temporal Difference error converges to 0. With regards to fear, remember that our model of fear is not parameterized by the risk-perception bias, i.e., fear=$V_-(s)$, *not* fear=-$V(s)$ if $V(s)<0$. This could be a wrong assumption and this can be tested. Currently we observed higher (and increasing) fear levels for outcome-control biased gambling reflecting the repeated exposure to the negative pay-off of gambling option B generating this $V_-(s)$ signal the next time the state just before B is visited. This predicts that individuals who suffer from gambling disorder in fact do experience fear prior to a potential loss, even though they don't use the negative predictions to update state values of states that are predictive of loss. If experimental data suggest that such individuals do not feel fear, this points towards the direction of the alternative fear model, i.e., fear= -$V(s)$ if $V(s)<0$. Our results thus predict that individuals who suffer from gambling disorder can experience fear (or worry) about near losses *and* distress due to the loss, but do not get distress signals just before the loss because their Temporal Difference error does not take into account the probability of losing, i.e., they do not learn based on the loss because they do not propagate back the loss signal to earlier states. This highlights that problem gambling is a behavioural and learning disorder [7], and is consistent with decreased (compared to normals) activation of the ventral striatum when problem gamblers are presented with gambling cues [7].

In the risky world, control bias results in higher joy at learning onset but eventually also in more distress. High onset joy is caused by overestimating positive outcomes in the Temporal Difference error, while high distress and high joy at learning convergence are caused by more extreme Temporal Difference errors caused by the random distribution of rewards. This high joy / high distress pattern at convergence is not seen in the realistic (no bias) case, because here the state values converge to weighted averages based on all possible next states. As such, in the realistic case, state values -- on average -- have smaller sized Temporal Difference errors than control-bias cases. This suggests that control-assuming individuals develop a preference for high-risk environments because they *experience high joy* when they encounter such environments, explaining the relation between risk-taking, sensation seeking, and perception of control.

In the lack of control task, we removed the ability to control actions in the standard T-maze. Agents learn as they would do but now select their actions randomly. This presents a situation in which we can investigate the effect of the assumption of control on value updates decoupled from what those values do to agents' behavioural policies. The results match with the general observation: control bias lowers the intensity of fear, lowers the intensity of distress, and increases the intensity of joy. Obviously there is no difference between action control and outcome control, as the T-maze does not have a biased coin ($B_1$/$B_2$). An interesting interpretation of this is the following. If capability (failure of action) is assumed to be modelled by randomness in action selection (i.e., I select $a_1$, but $a_2$ happened), then in the face of extreme action failure, optimistic agents have a much more pleasant experience even though this is unrelated to their performance. In other words: *optimism can help coping with failure*.

In an additional experiment we found that gambling behaviour is difficult to counter. Given sufficient exploration, an additional distracter or a pre-gambling punishment just before option B only delays the onset of gambling behaviour (Figure 3). Further, a high-stakes payoff structure for B ($p$(-2)=0.9 and $p$(17)=0.1) with the same expected return as the original gambling choice results in individuals that get addicted only when they receive the high B pay-off the first time as evidenced by the small proportion of individuals that pick option B more than 900 times per trial and the large proportion of individuals that almost never pick option B (Figure 3). High stakes make initial high pay-off a critical event in developing gambling disorder as proposed earlier by others [14].

Our results also indicate that there is a clear difference between "healthy" perception of control, i.e., the assumption that one has control over ones actions, and "unhealthy" perception of control, i.e., the assumption that one has control over outcomes of actions. When optimism is redefined in terms of a bias towards weighting future possibilities, it therefore seems useful to differentiate between *action optimism*, i.e., optimism as a bias towards assuming chosen actions will succeed which translates to one's ability to influence the world in general, versus *outcome optimism*, i.e., a bias towards assuming certain specific outcomes will follow particular actions which translates to one's ability to produce intended outcomes. Our simulations suggest that action optimism is positive to the individual: they tend to produce higher intensities of joy and lower levels of fear. In Reinforcement Learning this optimistic assumption is needed to learn optimal policies and is embodied in the *optimal value function* [60]. This is in line with optimism and hope theory, as mentioned earlier. Outcome optimism on the other hand can lead to risk taking and gambling behaviour, as shown in our simulation experiments. Interestingly this is really a different kind of optimism close to wishful thinking (the "submissive" version) and manipulative power (the "dominant" version), and can become a harmful trait to the individual. There is, however, a fine line between the two: outcome optimism is the natural extension of action optimism as in terms of probability biases this is simply the next step.

To find out how far away this next step could be, we further tested if the occurrence of gambling is specific for the extreme case of outcome optimism, i.e., a $MAX_{as}$ update function. We modified the Bellman update function to multiply the probability of an outcome *p(s')* based on $e^{v(s')}$. This weighing shifts the perceived probability of an outcome towards those outcomes with higher values. Results show (Figure 4) that in this case the occurrence of gambling behaviour is more dependent on the pay-off structure: agents with such a value function will slowly fall for the gambling scenario *if* the difference between the high frequency loss and low frequency gain is large enough for the exponential weighing to tip over the value function to the wrong side (Figure 4, mid). This is indicated by the decreasing reward over time and the fact that option B is chosen more often as shown in the distribution. This neither happens with action optimistic agents (Figure 4, bottom), nor with exponentially weighed agents in our standard gambling scenario (Figure 4, top).

Our results have to be put in perspective. It is unknown how this model of risk taking reacts to other factors than the ones studied here. For example, what are the roles of near-miss effects? How sensitive is the model to learning parameters such as discount and learning rate? What is the influence of different levels of optimism on the exploration of the state space? And, what if the state space is non-Markovian? These questions are also relevant for modelling emotion in relation to RL. Some of these issues have been addressed in previous work [61], for example that variation of $MAX_a$ (high-hope) versus Bellman (low-hope) updates replicates the finding that high-hope individuals perform better than low-hope individuals (and have lower fear because *V(s)* will reflect only positive actions). The reason is that the high-hope individuals explore differently: they selectively cut off maze arms that lead to punishment, while the paths in the maze that lead to positive outcomes are not impacted by the punishment. Investigating these issues is not an easy task and doing so would distract from the current article's focus: to model risk taking variation in the Temporal Difference error and then test if this variation replicates behavioural and affective dynamics in a selected variety of task involving different forms of uncertainty. These important issues are in line with our future research.

**Conclusion**

We modelled the effect of optimism as an exaggeration of probabilities for either positive actions or positive outcomes of actions in the calculation of the Temporal Difference error. This straightforward manipulation of risk perception sheds light on the relation between fear, sensation-seeking, risk-taking behaviour, gambling disorder, and (perception of) control. We show that high intensity of sensations and low levels of fear co-occur due to optimistic risk perception. We found that *overly* optimistic risk perception (outcome optimism) results in risk taking and in persistent gambling behaviour in addition to high intensity of sensations. Our results suggest that there is a clear difference between action optimism, i.e., the optimistic assumption that one has control over ones actions, and outcome optimism, i.e., the overly optimistic assumption that one has control over specific outcomes of actions.

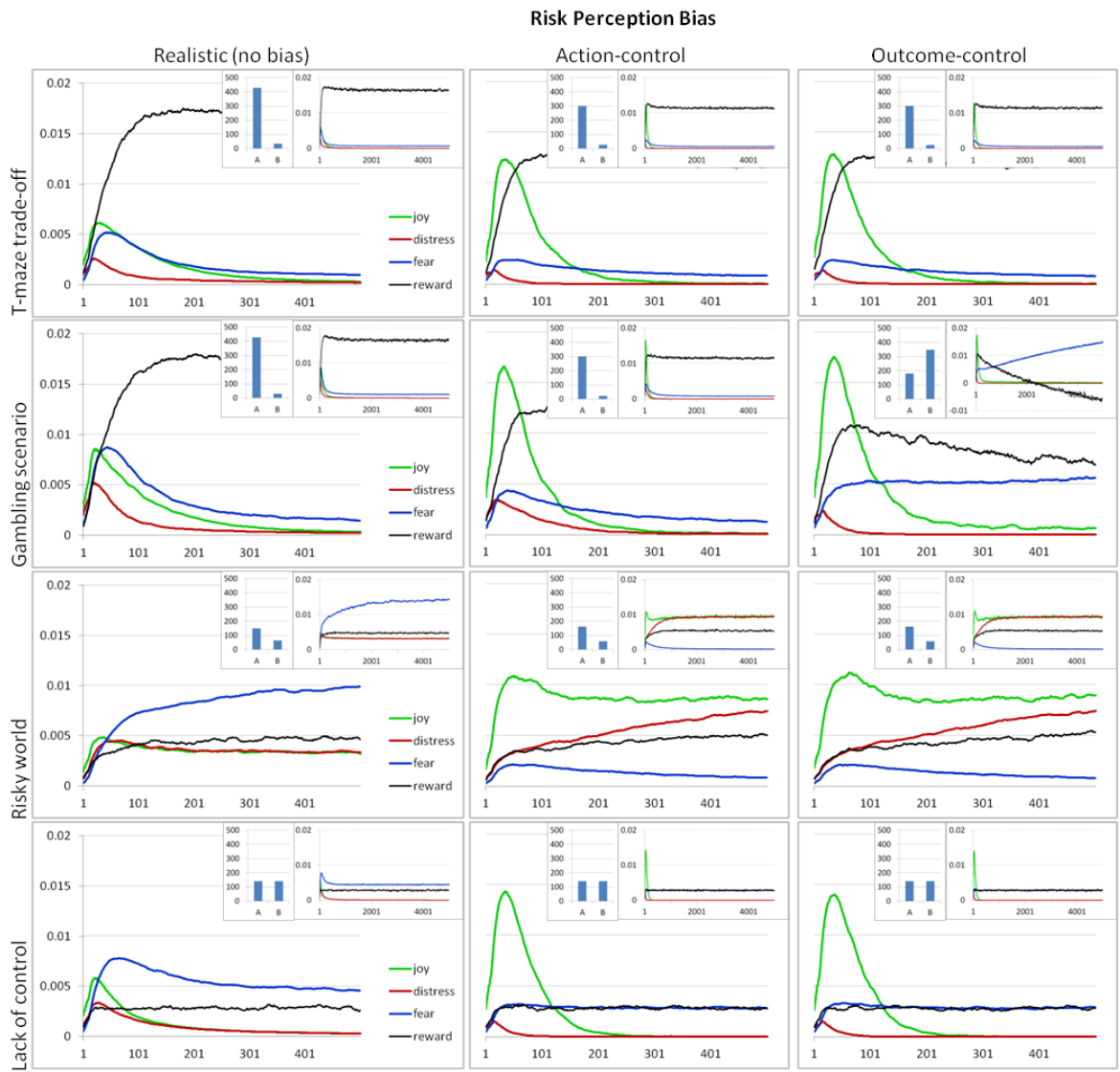

*Figure 2. Population plots over time (steps since start of learning at the x-axes) of the intensity of* joy (TD$_+$ error)*, distress (TD$_-$ error), fear (V$_-$(s))* and reward *(y-axes) for variations of risk perception bias in the Temporal Difference error and different tasks. Main plots show averages over 5000 individuals for the first 500 steps (learning onset). Insets show the convergence of these plots after 5000 actions, and the average number of times an agent picks A and B (averaged over 5000 individuals). See text for details.*

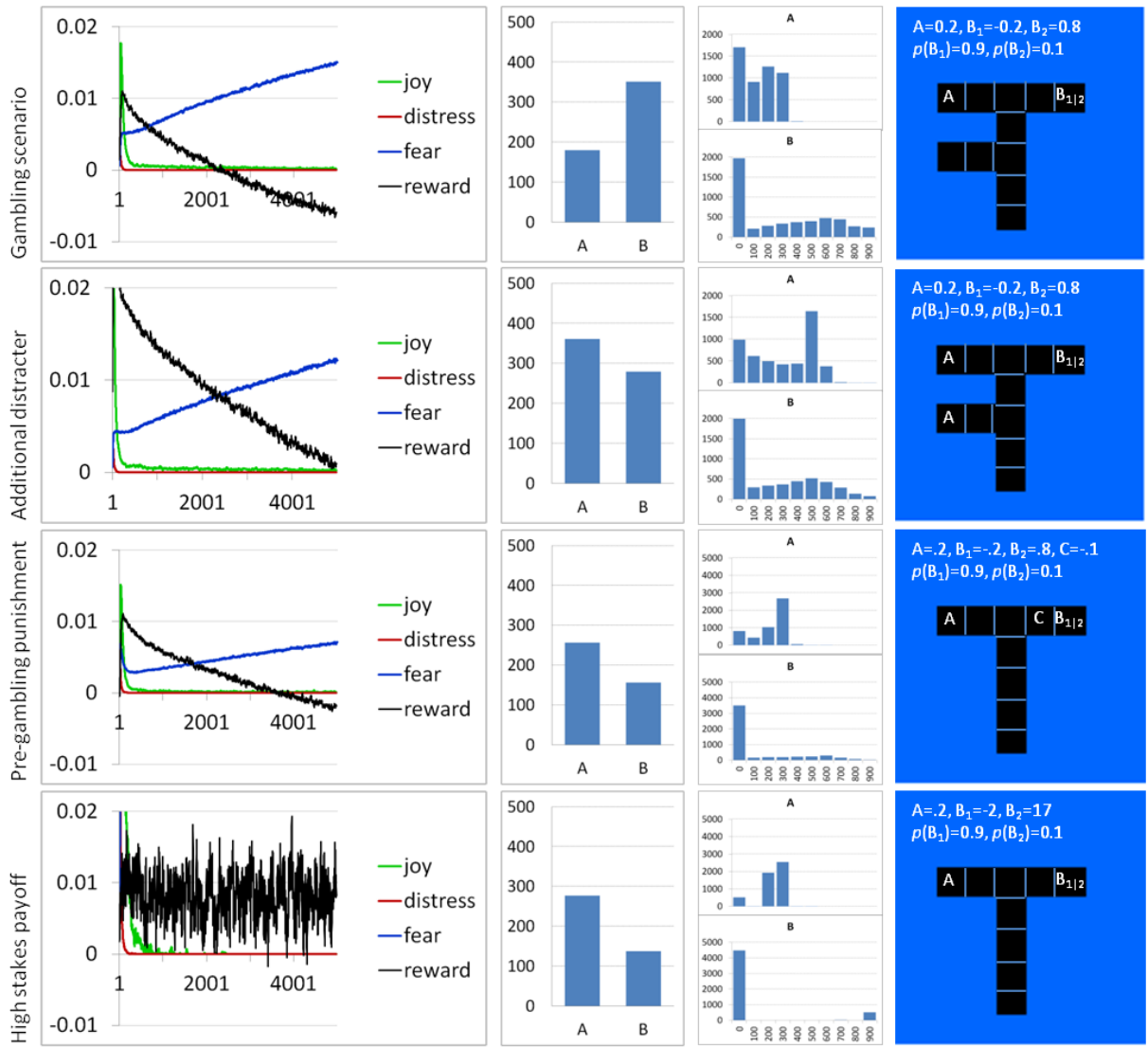

Figure 3. The gambling scenario (top row) with effects on gambling behaviour modulated by a second distracter in the environment (second row), a small punishment just before gambling option B (third row), and 10x higher stakes (bottom row). As in figure 2, plots and distributions are based on aggregation of the sample data (n=5000). From left to right: convergence plots (steps on the x-ases, intensity of joy ($TD_+$ error), distress ($TD_-$ error), fear ($V_-(s)$) and reward on the y-axes), average number of times an agent picks A and B over all agents (averaged over 5000 individuals), and the distribution for picking A and B for all 5000 agents showing how many agents (y-axis) pick A and B how often (x-axis) (for example, in the high stakes payoff, almost no agents pick the gambling option B, but those who do pick B and get the high pay-off the first time are hooked). See text for details.

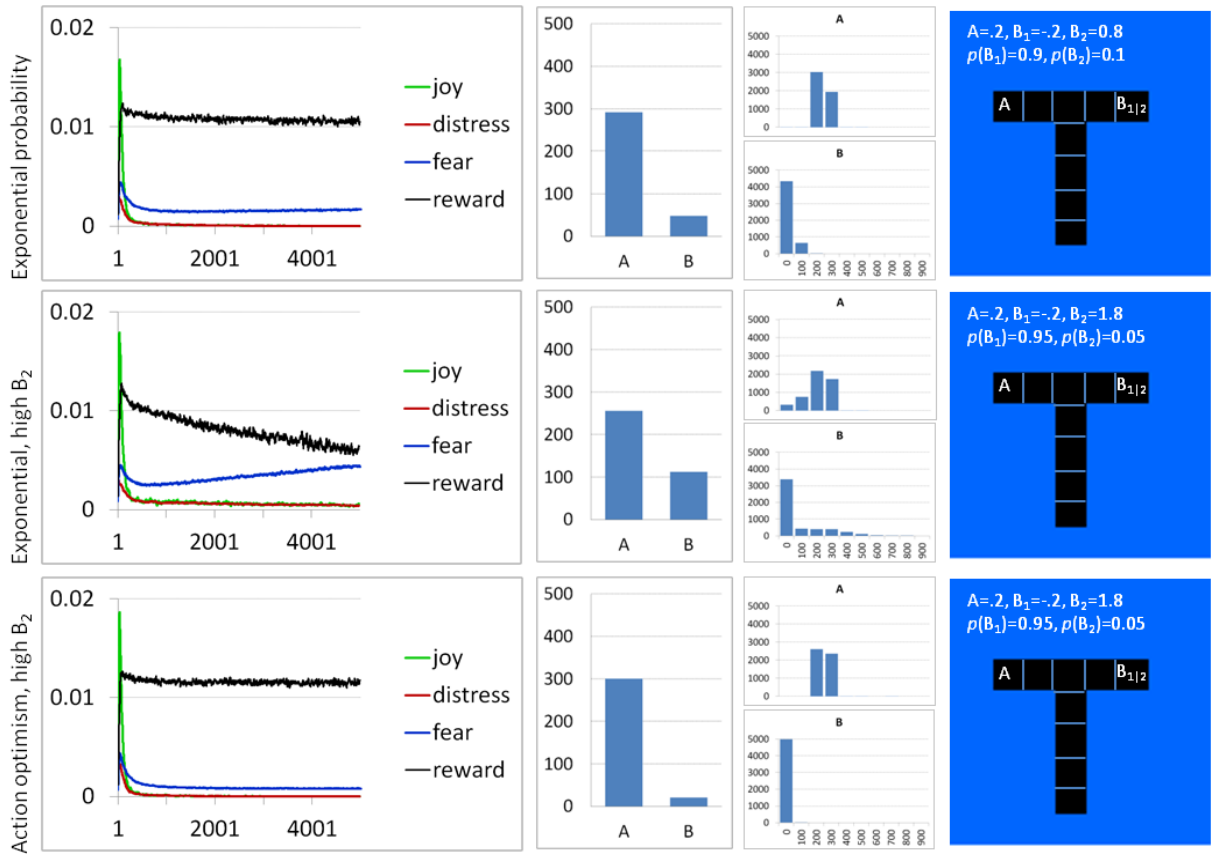

*Figure 4. The gambling scenario with effects on gambling behaviour induced by an outcome-value dependent exponential weighing of outcome probability (top row), a combination of this exponential weighing and a different pay-off structure with a higher but rarer pay-off (mid row), and action optimistic agents ($MAX_a$) in this new pay-off scenario. The new pay-off structure is on average the same as the old one. The figures from left to right are the same type as those explained in Figure 3.*